\documentclass{article}

\usepackage[nonatbib, final]{nips_2016}

\usepackage{nips_2016}
\usepackage{enumitem}
\usepackage[utf8]{inputenc} 
\usepackage[T1]{fontenc}    
\usepackage{hyperref}       
\usepackage{url}            
\usepackage{booktabs}       
\usepackage{amsfonts}       
\usepackage{nicefrac}       
\usepackage{microtype}      
\usepackage{amsmath}
\usepackage{environ}

\usepackage[pdftex]{graphicx}
\usepackage[font=footnotesize]{caption}
\usepackage[font=footnotesize]{subcaption}

\usepackage{tikz}
\usetikzlibrary{shapes,arrows}
\usepackage{amsmath,bm,times}

\usetikzlibrary{positioning}

\title{Siamese Regression Networks with Efficient mid-level Feature Extraction for 3D Object Pose Estimation}

\author{Andreas Doumanoglou, Vassileios Balntas, Rigas Kouskouridas and Tae-Kyun Kim
\thanks{All authors are with the Imperial Computer Vision and Learning Lab (ICVL), at the Department of Electrical and Electronic Engineering, Imperial College London, UK, {\tt\small \{a.doumanoglou12, v.balntas15, r.kouskouridas, tk.kim\}@imperial.ac.uk}}%
}
\begin{document}

\maketitle

\begin{abstract}
In this paper we tackle the problem of estimating the 3D pose of object instances, using convolutional neural networks. State of the art methods usually solve the challenging problem of regression in angle space indirectly, focusing on learning discriminative features that are later fed into a separate architecture for 3D pose estimation. In contrast, we propose an end-to-end learning framework for directly regressing object poses by exploiting Siamese Networks. For a given image pair, we enforce  a similarity measure between the representation of the sample images in the feature and pose space respectively, that is shown to boost regression performance. Furthermore, we argue that our pose-guided feature learning using our Siamese Regression Network generates more discriminative features that outperform the state of the art. Last, our feature learning formulation provides the ability of learning features that can perform under severe occlusions, demonstrating high performance on our novel hand-object dataset.
\end{abstract}


\section{Introduction}\label{Sec_Intro}
Detecting objects and estimating their 3D pose is one of the most challenging tasks in computer vision, since severe occlusions, background clutter or large scale changes, dramatically affect the performance of any contemporary solution. State of the art methods make use of Hough Forests for casting patch votes in the 3D space \cite{tejani2014latent,andoum2016recovering} or train CNNs to either perform classification into the quantized 3D space \cite{johns2016pairwise} or learn features that are later fed to a Nearest Neighbor scheme for 3D object pose template matching \cite{wohlhart2015learning}. The lack of research on Deep Networks for angle regression prove that directly regressing object poses in angle space is not trivial, with the objective function appearing to have many local minima.  Albeit the recent advances in classification tasks using CNNs, a framework that is able to perform direct regression in angles, while jointly learning discriminative features has yet to be built.  

Towards this end, in this paper, contrary to the state of the art methods, we are interested in learning an end-to-end framework that directly regresses object pose angles. Recent works \cite{sun2014deep,hoffer2015deep} demonstrated successful results by using siamese networks, which can improve the network learning capabilities by exploiting additional information about the relationship between the training samples. Inspired by this, we present the Siamese Regression Network that enforces a relationship between feature and pose space by applying a novel loss function, which can boost the performance of a regression network layer. Thus, this network is able to perform single-shot end-to-end regression for object pose estimation, without requiring pairs of inputs at testing time. Apart from that, we experimentally evaluate the effect of some other factors that play an important role in successful regression such as feature normalization and batch formation. Our Siamese Regression Network, on the other hand, proved to learn more discriminative features optimized for our particular problem, as compared to a state of the art feature learning technique using CNNs \cite{wohlhart2015learning}. Finally, we are interested in handling severe occlusions in the object pose estimation task, a particularly interesting problem constantly arising in real-life applications. However, estimating accurate pose when a significant portion of the object is missing is a very challenging task which drastically degrades the performance of previous arts \cite{wohlhart2015learning,hinterstoisser2012accv,brachmann2014learning,lim2014fpm,wu20153d,mottaghi_cvpr15}. We show how our loss function can be easily modified to handle cases of severe occlusions. To evaluate our regressor on such cases, we built our own challenging dataset which demonstrates an object being manipulated by a human hand. Results show that our method can very well handle severe occlusion, reaching accuracy levels of non-occluded objects.

In summary our paper offers the following contributions:
\begin{itemize}[topsep=0pt,itemsep=-1ex,partopsep=1ex,parsep=1ex,leftmargin=3ex]
\item We present Siamese Regression Network which, to the best of our knowledge, is the first CNN-based framework for regressing object poses in angle space.

\item We boost the performance of our system by introducing a novel loss function for feature-guided pose regression.

\item In turn, we show that pose-guided feature learning results in more discriminative features than the ones of \cite{wohlhart2015learning} and are, experimentally proven, optimized for the particular task of 3D object pose estimation.

\item We show how our loss function can be adapted to deal with severe occlusions and evaluate our system on a new challenging dataset containing an object captured under severe occlusions. Furthermore, experimental evaluation on a benchmark dataset  \cite{hinterstoisser2012accv} provide evidence of our system outperforming the state of the art.
\end{itemize}

The remainder of the paper is organized as follows. In Section \ref{Sec_Rel} we provide an overview of the related work, while our proposed approach is introduced in Section \ref{Sec_Meth}. In Section \ref{Sec_Exp} we present evaluation of our method compared to the state-of-the-art on two datasets. Finally, in Section \ref{Sec_Con} we conclude with final remarks and an outlook to future work.


\section{Related Work}\label{Sec_Rel}
Recognizing and detecting objects along with estimating their 3D pose has received a lot of attention in the literature. Early works made use of pointclouds  to facilitate Point-to-Point matching \cite{drost2010model,rusu2009fast}, while the advent of low-cost depth sensors \cite{hinterstoisser2012accv,rios2013discriminatively} provided additional data in favor of textureless objects. Hinterstoisser et al. designed a powerful holistic template matching method (LINEMOD) based on RGB-D data that suffers in cases of occlusions. Tejani et al. \cite{tejani2014latent} integrated LINEMOD into Hough Forests to tackle the problem of occlusions and clutter. The work of Brachmann et al. \cite{brachmann2014learning} along with its recent extension to RGB-only images \cite{brachmann2016uncertainty} employ a new representation framework that jointly maps 3D object coordinates and class labels. Hodan et al. \cite{hodan2015detection} presented a method that tackles the complexity of sliding window approaches via a fast-filtering technique followed by a voting procedure for hypotheses generation, while fine 3D pose estimation is performed via a stochastic, population-based optimization scheme. In turn, in \cite{song2014sliding} exemplar SVMs are slided in the 3D space to perform object pose classification based on depth images.

Deep learning has only recently found application to the 3D object pose estimation problem. Doumanoglou et al. \cite{andoum2016recovering} suggested using a network of stacked sparse autoencoders to automatically learn features in an unsupervised  manner that are fed to Hough Forests for 6D object pose recovery and next-best-view estimation. In \cite{johns2016pairwise} Johns et al. employed a CNN-based end-to-end learning framework for classification of object poses in the 3D space and next-best-view prediction. In turn, in  \cite{crivellaro2015novel} a CNN was used to learn projections of 3D control points  for accurate 3D object tracking, while in \cite{krull2015learning} a CNN is utilized in a probabilistic framework to perform analysis-by-synthesis as a final refinement step for object pose estimation. In \cite{wohlhart2015learning} 3D pose estimation is performed by a scalable Nearest Neighbor method on discriminative feature descriptors learned by a CNN. 

To the best of our knowledge, this paper presents the first CNN-based framework for regressing object poses in the continuous 3D space. The work of Kendall et al. \cite{kendall2015posenet} regresses camera poses in the continuous 3D space\footnote{camera pose estimation is the inverse of object pose estimation} but does  not offer any end-to-end learning since it makes use of the pretrained VGG network adding just a softmax layer for regressing the four quaternions of the camera pose. The method of Sun et al. \cite{sun2014deep} offers a learning framework that learns a new face representation by joint identification-verification. As far as feature learning for 3D object pose estimation is concerned, our work shares similar ideas with the method of Wohlhart et al. \cite{wohlhart2015learning} that learns feature descriptors with pairs and triplets. However, we argue that our learned features are pose-guided and as experiments prove, more discriminative, which in fact suggests that they are optimized for the particular task of 3D object pose estimation.

\section{Siamese Regression Network}\label{Sec_Meth}

\subsection{Object Pose Estimation Using Regression CNN}

We first formulate the problem of object pose estimation as a regression problem. Let $ x \in \mathbb{R}^{W\times H \times 4}$ be an RGBD (4 channels) image depicting a centered object having width $W$ and height $H$. Pose estimation is the problem of learning a regressor $g:\mathbb{R}^{W\times H \times 4} \to \mathbb{R}^{M}$, where $M$ is the dimensionality of the pose representation used. For example, Euler angles require $3$ angles to be defined ($M=3$) whereas quaternions suggest $M=4$.
Regressing euler angles directly can be problematic due to multiple
problems such as periodicity \cite{yi2015learning}, and the non-continuous
nature of the euler angle space \cite{kendall2015posenet}. For
example, poses that are very
similar visually might be far away in euler angle space, making regression harder. Therefore,
similar to previous work \cite{kendall2015posenet} on regressing camera pose, we also use the quaternion
representation, which does not suffer from the same
problems. 

For the task of estimating the regression function $g$ we train a convolutional neural network (CNN). We use the simple architecture similar to \cite{wohlhart2015learning} that consists of 2 convolutional layers, and 2 fully connected layers (we have removed the max-pooling layers as we saw that they slightly degraded the performance). On top of that, we added another fully connected layer that outputs $M$ units to estimate the object pose. If we consider the layer just before the last regression layer as the features learned by the network, we can describe the output of the our network as:

\begin{equation}
  \label{eq:cnn}
  p = g(f(x))
\end{equation}

where $f(x)$ is the output of the feature layer, $g$ the regression layer function and $p$ is a pose vector returned by the network for the input image ${x}$.
Given a training set that contains combinations of training samples of the form $\{x_i, y_i \}$, the most commonly used method of training a regression network is by minimizing the Mean Square Error (MSE) between the estimation $g(f(x_i))$ and the ground truth $y_i$ and back-propagating the error. If we split the training set into mini batches of $K$ samples each, the regression loss can be written as:

\begin{equation}
  \label{eq:regression}
  \ell_R = \sum_{n=1}^{K} ||g({f(x_{n})}) - {y_{n}} ||_{2}^{2}
\end{equation}

\tikzstyle{feats} = [draw, fill=blue!15, rectangle, 
    minimum height=3em, minimum width=3em]
\tikzstyle{pose} = [draw, fill=teal!30, rectangle, 
    minimum height=3em, minimum width=1em]
\tikzstyle{loss} = [draw, fill=cyan!20, rectangle, 
    minimum height=1em, minimum width=1em]

\begin{figure}
\centering
\includegraphics[width=0.8\textwidth]{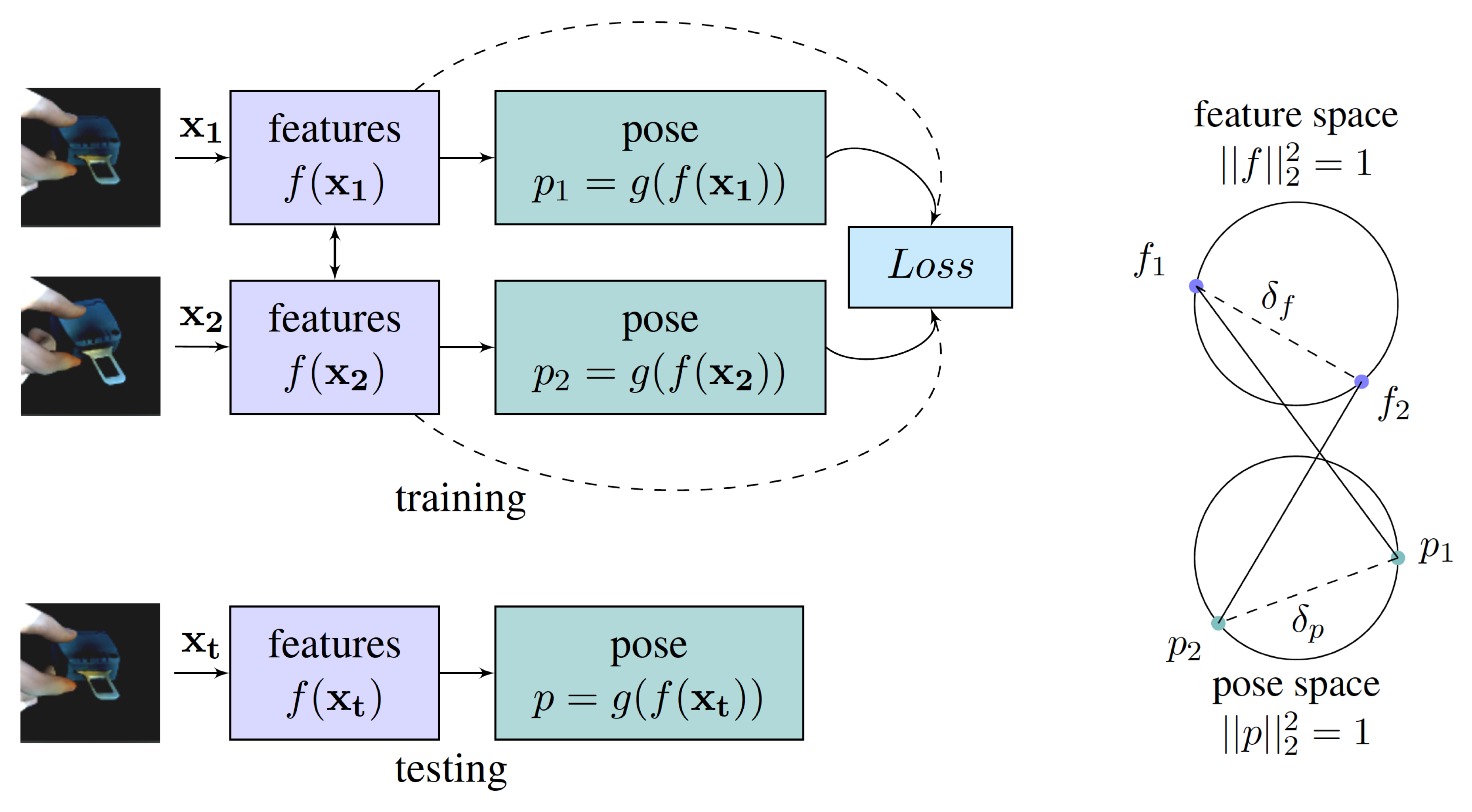}
\caption{(left) Our training and testing architectures. We enforce a siamese architecture for regressing relative distance between feature and pose spaces. During testing, we extract a branch of the network, and use it for regression. (right) Illustration of our feature-guided pose regression loss. The loss seeks to associate distances in the $L2$ normalised feature space with the $L2$ normalised pose space. }
\label{fig:idea}
 \end{figure}

\subsection{Siamese Regression Objective}

Previous work has shown that the feature layer $f$ is able to learn representations that can be successfully applied in nearest neighbour matching \cite{wohlhart2015learning} or face identification - verification \cite{sun2014deep}. However, end-to-end regression learning with CNN in angle space proved to be a very challenging task, with researchers resorting to indirect solutions, such as Nearest Neighbor template matching \cite{wohlhart2015learning} or ad-hoc angle estimation methods like $arctan$ \cite{yi2015learning}. Therefore, inspired by \cite{sun2014deep}, we want to enhance the feature learning process by using additional information available during training, in order help the end-to-end regressor converge to a better minimum. Thus, our goal is to enforce a loss function in the feature layer $f$, in a way that the learned features are more appropriate and useful for the regression task in the last layer $g$.

In order to enforce a second loss function in this layer, we utilize the siamese architecture that has been very successful for learning non-linear feature embeddings with convolutional neural networks \cite{hadsell2006dimensionality}.  The siamese architecture consists of two (or more) branches of the same CNN that share weights and encode two inputs processed in parallel. Subsequently, a loss function can be introduced based on both outputs, which makes it possible to compare different samples of our training data passing through our network in a meaningful way. 

Our study on the regression problem concluded that there is a relation between the feature and the angle space which helps a regression network layer perform much better. The relationship is the following: the euclidean distance between two sample images represented in feature space, should be maintained the same with the distance between the same samples as represented in angle space, during training. Fig. \ref{fig:idea} shows an illustration of our idea. In order to enforce such relation we use a siamese network to pass through the network pairs of samples and apply an objective term on them. The pairs have the form: $\{ {x_1}, {y_1}, {x_2}, {y_2}\}$ where $x$ represent the raw input, and $y$ the pose vector ground truth. We enforce the following loss function for feature guided regression

\begin{equation}
  \label{eq:loss_feat}
    \ell_F = \sum_{n=1}^{K} ||f({x_{n,1}}) - f({x_{n,2}}) ||_{2}^{2} -  ||{y_{n,1}} - {y_{n,2}} ||_{2}^{2}
\end{equation}

Intuitively, minimizing this loss enforces the $L2$ distance between the features in the sample pair, to be close to the $L2$ distance between the ground truth of their poses. In order to avoid weighting any of the above parts of the objective loss term, we normalize the output of the feature layer as well as the output of the pose layer to have unit norm (if using quaternions as pose representation, they already have unit norm). In fact, as we will show in experiments, this normalization has a positive effect in training angle regression even without using our extra feature term.

It should be mentioned that the siamese network is only used during training to help the regression task. During testing, only a single image produces a pose estimation without the need of providing a pair for the image.

\subsection{Feature Guided Pose Regression}

Combining the regression loss with the feature loss, we get 

\begin{equation}
  \label{eq:loss_feat}
    \mathcal{L} = \ell_R + \ell_F   + \lambda \cdot ||w||_{2}^{2}
\end{equation}

where $\lambda \cdot ||w||_{2}^{2}$ is a term to regularise the
weights of the convolutional neural network. By enforcing this loss in
the proposed siamese regression network, we are able to
simultaneously focus on both features that are able to work well in a
nearest neighbour framework, and on the fully connected last layer
that regresses the poses directly. Indeed, in our experiments we show
that enforcing the feature term in the loss leads to better pose
estimation in the final layer. 

In Table \ref{tab:learn}, we describe the relationship between the two
loss functions $\ell_R \text{ and } \ell_F$, and the parts of the CNN
weights that are updated. We note that the weights related to the
feature learning, are updated using information from both losses,
while the weights related with the pose regression, are only updated
based on the $\ell_R$ loss.


\subsection{Pose Guided Feature Learning}
Despite the fact that the loss function of from Eq.~\ref{eq:loss_feat}
mainly aims to learn a better regressor for the pose in the
final layer $p=g(f(x))$ of the convolutional network, it can be argued
that the features that are learned in the $f(x)$ layer of the network
can be more discriminative.

Previous work on 3D feature learning with siamese networks has
focused on optimising the feature embeddings using triplets. Triplet training
samples contain an anchor, a positive sample and a negative
sample. The authors from \cite{wohlhart2015learning} form the triplet by using two close views of the
object as anchor and positive samples, and a view with significantly
different pose as the negative. What they try to optimize is the anchor and the positive sample to be closer in feature space than the anchor and the negative one. They also use pairs of images of similar pose but different appearance and try to minimize their distance in feature space in order to learn features immune to different lightning conditions and noise.


On the other hand, our loss focuses on forcing the feature distance
between a pair to be equivalent to the pose distance. Thus, it is more
appropriate for a nearest neighbour framework, since the features are
optimized to be relative to the pose distance.  Indeed, in our experiments
show that enforcing our loss from Eq.~\ref{eq:loss_feat} results in
features that are more suited for nearest neighbour matching. 

We should note that using the objective function of \cite{wohlhart2015learning} instead of $\ell_F$ in order to help regression didn't work, with the network showing similar convergence behavior of the simple regression without using the extra term. This is a clue that our objective does indeed help regression, while at the same time regression helps building more discriminative features appropriate for pose estimation.

\begin{table*}[t]
\caption{Our learning algorithm.}
\label{tab:learn}
\begin{center}
\begin{tabular}{p{350pt}}
\toprule
\textbf{input}: training set $\mathcal{X}$, CNN feature parameters $w_F$, CNN pose parameters $w_R$, learning rate $\eta$\\
\midrule
\textbf{for} epoch e=1:N \textbf{do} \\
\quad sample M mini-batches from $\mathcal{X}$ using pairs with both similar and different poses\\
\quad \textbf{for} mini-batch b=1:M \textbf{do} \\
\large\quad \quad$\nabla w_{F}=\frac{\partial \ell_F}{\partial w_F} + \frac{\partial \ell_R}{\partial w_F}$ \\
\large\quad \quad$\nabla w_{R} = \frac{\partial \ell_R}{\partial w_R}$\\
\quad \quad update $w_{R}=w_R-\eta(e)\cdot w_{R}$ \\
\quad \quad update $w_{F}=w_F-\eta(e)\cdot w_{F}$ \\
\quad \textbf{end} \\
\textbf{end} \\
\textbf{output} $w_R,w_F$ \\
\bottomrule
\end{tabular}
\end{center}
\end{table*}

\subsection{Siamese Pair Sampling}
Considering a dataset of $M$ training samples of the form
$\{\bf x_i, \bf y_i \}$, there exist $\binom{M}{2}$ possible pairs to
be used in the siamese training process described above. Since the
number of pairs can become very large, several authors explored different techniques of sampling 
or mining hard negative pairs \cite{wang2014learning,simo2015discriminative}.

Although we do not explicitly have positive and negative pairs since
the training process is done in the same object, we approximate such
pairs by spliting a batch of size $K$ between pairs that are both close in
the pose space, or have very large pose differences. We
examined that random choice of pairs in terms of their pose
difference perform inferior to our formation. Interestingly, forming batches using both similar and different pose pairs, is a factor that improves regression on its own, even without enforcing any constraints on such pairs. In the experiments we will show the relative performance gain of using well formed batches compared to simple regression and enforcing our sample pair objective.

\subsection{Handling occlusions}
Tackling occlusions, that is estimating the object pose when a major part of the object is missing or occluded, requires features that are robust in such conditions and one should explicitly enforce this property. We note that the form of Eq.~\ref{eq:loss_feat} makes it convenient to support building such features: if we generate training samples with the object occluded, and using its annotation render a \textit{clean} object having the same pose, we can enforce a similar term between the occluded and the clean images:

\begin{equation}
  \label{eq:occl}
   \ell_{oc} = ||f(x_{occluded}) - f(x_{clean})||^{2}_{2} - ||y_{occluded} - y_{clean}||^{2}_{2}
\end{equation}
 
where $x_{occluded}$ and $x_{clean}$ are images depicting occluded and
clean objects respectively. Fig.~\ref{fig:dataset} in the experiments section shows examples of such images. This term can be added to the $\mathcal{L}$ loss in order to tackle the severe occlusion problem.


\section{Experiments}\label{Sec_Exp}

\begin{figure}[t]
  \hspace{-0.2in}      
  \begin{subfigure}[t]{0.32\linewidth}
              \includegraphics[width=47mm]{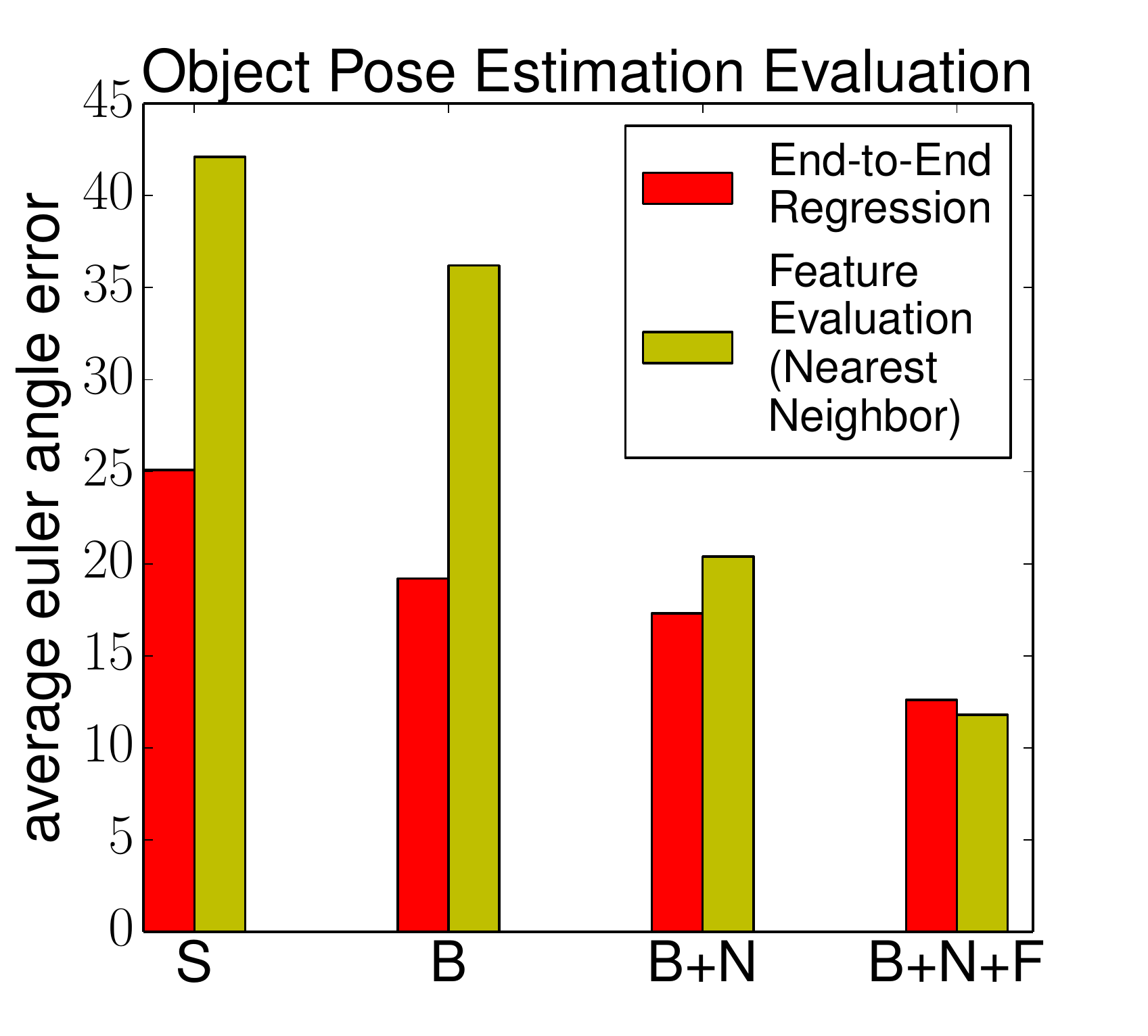}
              \caption{}
              \label{fig:selfpose}
	\end{subfigure}
\hspace{-0.1in}
\begin{subfigure}[t]{0.32\linewidth}
		\includegraphics[width=55mm]{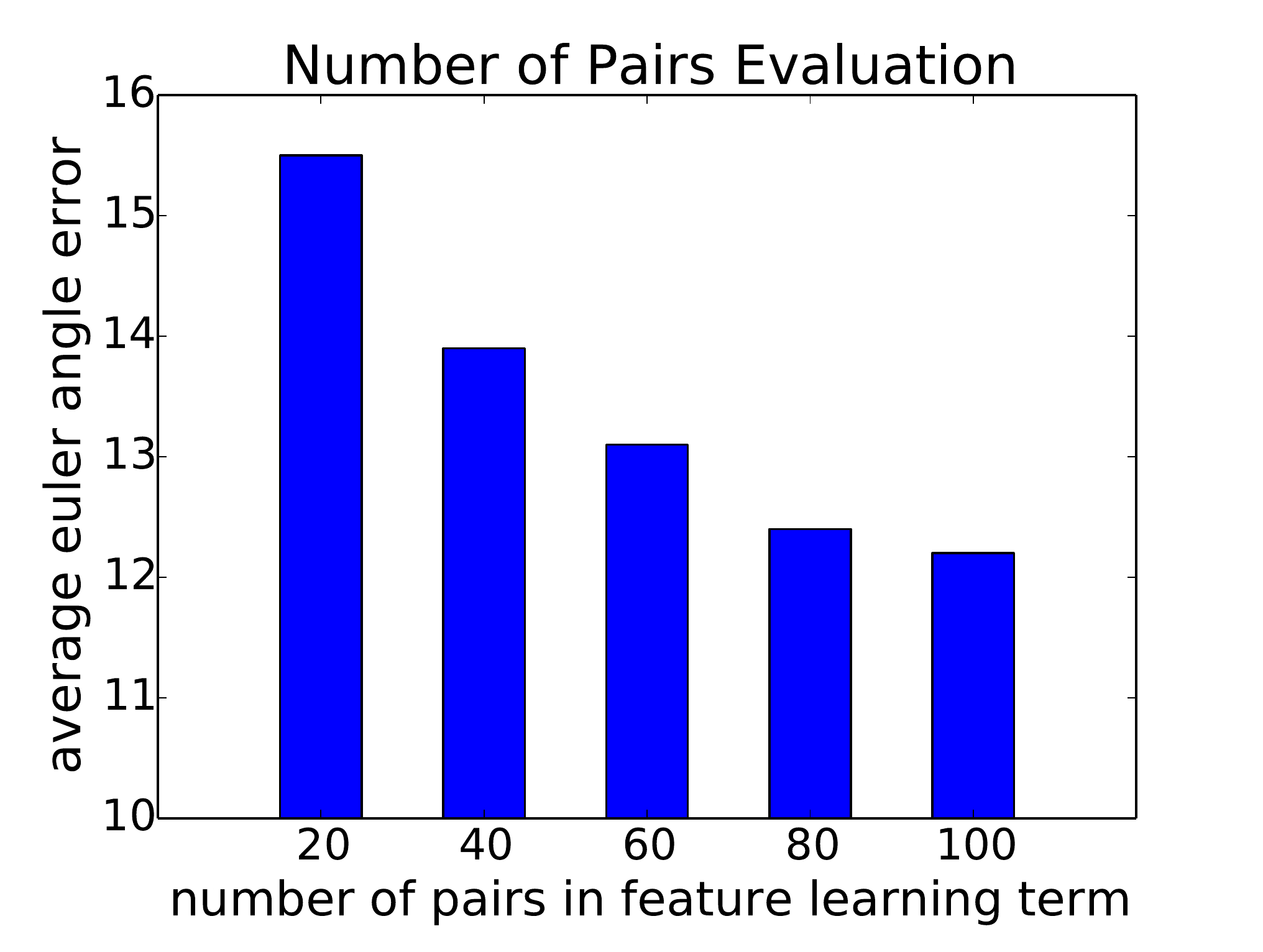}
                \caption{}
                \label{fig:selfpairs}
	\end{subfigure}
\hspace{0.2in}
	\begin{subfigure}[t]{0.32\linewidth}
		\includegraphics[width=55mm]{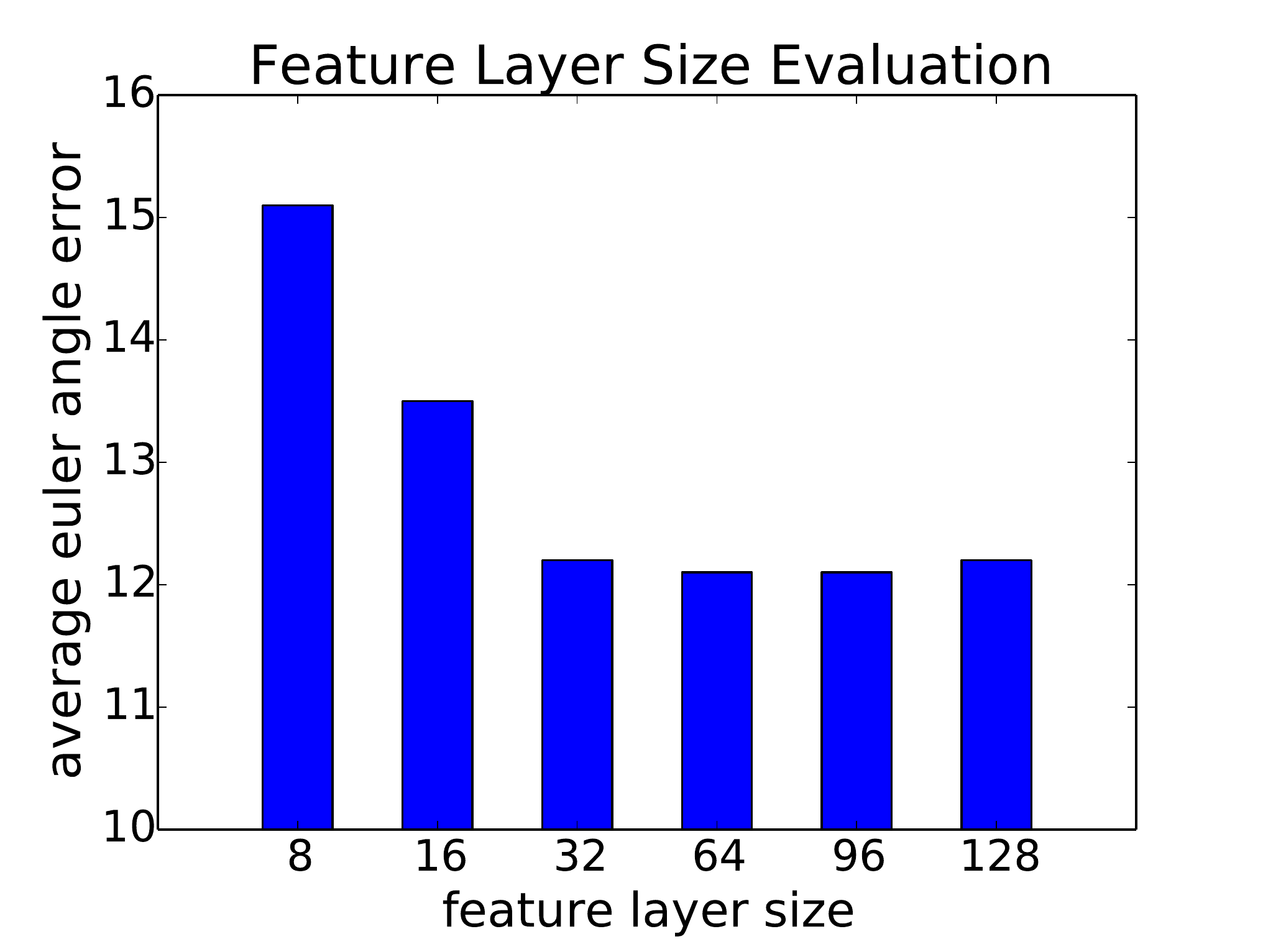}
                \caption{}
                \label{fig:selffsize}
	\end{subfigure}
	\caption{a) End-to-end regression evaluation as compared to our learned features using nearest neighbor on various network configurations: S is simple regression, B is again simple regression with properly formed batch, N means that the network contains normalization layer after the feature and final layer, and F means that the network is trained using our new feature-guided pose regression objective. b) Evaluation of different number of pairs compared inside a mini-batch. c) Evaluation of the length of the feature layer.}
	\label{fig:self_eval}
\end{figure}

Our convolutional regressor is a simple convolutional neural network with similar architecture to \cite{wohlhart2015learning} that has the following architecture: \{$Input(4,W,H)-Conv(16,8,8)-Conv(7,5,5)-FC(256)-FC(D)-FC(4)$\} where $Conv(N,K,K)$ represents a convolutional layer with $N$ filters of size $K\times K$, and $FC(D)$ a fully connected layer with $D$ outputs. Note that the feature layer $FC(D)$ is of variable length, something that allows a trade-off between feature extraction size and performance. We use $ReLU$ as the non-linearity in all our convolutional and fully connected layers apart from the last layer that produces the pose estimation where we used $tanh$. For training the network we use the stochastic gradient method~\cite{bottou2012stochastic}, with $0.9$ momentum and initial learning rate of $0.01$. We also decay the learning rate in each epoch, to avoid oscillations around local minima. 

In order to evaluate our method we used two datasets. The first one, which is also used for our parameter analysis, is the one of LINEMOD \cite{hinterstoisser2012accv}. More specifically, we worked with a variant of the RGBD images as used by  \cite{wohlhart2015learning}  where the objects are centered in the image so that no localization is required. This dataset contains about 3000 training images and 1000 test images per object. 

The second dataset is constructed by us and depicts a small object (a car belt) being manipulated by a human hand as seen in Fig. \ref{fig:dataset}. The focus of such dataset is to introduce realistic occlusions that are severe and have stronger effect on the learning process than using different object instances or types. In order to construct such dataset, we recorded an RGBD video, using Asus Xtion, of a human manipulating the object, and used a particle swarm optimization tracker \cite{tompson2014real,oikonomidis2011efficient} to track both the hand pose as well as the object pose. Having such information we can easily generate our needed pairs for our occlusion term (eq. \ref{eq:occl}). Such scenario appears in autonomous learning of object manipulation by robots, where the task is being demonstrated by a human. This dataset is very challenging since human hand introduces high level of occlusion which can significantly degrade the accuracy of pose estimation. Moreover, our dataset is larger, with about 21000 training and 5000 testing images.

Regarding the evaluation metric, we use the average Euler angle error, which is the average of the absolute difference in angle (in degrees) between the estimated pose and the ground truth, measured regarding the three principal axes. Such metric is more appropriate for our regression task and matches the one used in \cite{wohlhart2015learning} . Since we are using quaternions, we transform them in Euler angles after the estimation in order to perform the comparisons.

In the following subsections, we first evaluate different parameters of our network showing the relevant importance of each of our contributions. We also compare our siamese regression network with some base-line and state of the art methods showing the superiority of our method.

\subsection{Parameter Evaluation}

Fig. \ref{fig:selfpose} shows the evaluation of the different parts of our network, starting from a simple regression network and gradually adding elements of our siamese regression network. We show both the performance of our end-to-end pose estimation and the performance of our produced features using nearest neighbor template matching. The simple regression network performs worse, and the performance gradually increases by just using a better formed batch, then the normalization layers and the best performance is achieved when adding our feature learning term in the objective function. Interestingly, we notice that the our features using nearest neighbor exhibit similar behavior, but the increase in performance is more significant. When using our feature learning term, it slightly outperforms the end-to-end regression. As we will see in the next subsection, the regression is more affected by overfitting regarding the small size of the linemod dataset we used for the analysis.

We also experiment with the amount of pairs required for our feature term. Fig. \ref{fig:selfpairs} shows the regression performance for different values of the amount of pairs used. Using a batch that contains 300 training samples we see that the more pairs we use, the better the performance. However above 100 pairs we did not get any further significant improvement.

Last, we evaluated our network on different feature sizes, shown in Fig. \ref{fig:selffsize}. Again we see that the more features used the better the performance achieved. Above the size of 32 however, there is not significant improvement, which is in par with what was reported in \cite{wohlhart2015learning} .

\subsection{State Of The Art Comparisons}

Last, we performed a final evaluation of our siamese regression network compared with the method of \cite{wohlhart2015learning} , which is the most relevant work to ours and directly comparable. This work uses triplets and pairs formed by the training samples and learn to minimize an objective function using a convolutional neural network. This objective only tries to increase the euclidean distance in feature space of dissimilar samples, while enforcing similar samples to be close. The idea behind this is to build a mapping appropriate for nearest neighbor matching with some templates. 

Results are shown in Table \ref{soa-table}. We see that both the end-to-end regression and our learned features outperform the previous work. One reason for this is that the objective used by \cite{wohlhart2015learning}  does not take into account the actual task objective which is the pose estimation. Our method has as ultimate goal to learn the object pose directly, and therefore constructing more appropriate features for this task. On the other hand, we see that on the small dataset of \cite{hinterstoisser2012accv}, nearest neighbor performs slightly better than the end-to-end regression, which is prone to overfitting regarding the dataset size. When experimenting on our larger occlusion dataset, we see that the end-to-end regression is able to converge to a better minimum. It is clear that both our features and regression significantly outperform \cite{wohlhart2015learning}. Furthermore, our formulation gives us another opportunity to further improve the performance when we can generate synthetically the occluded and the clean image of an object. We see that by using equation \ref{eq:occl} pose error decreases even further, reaching accuracy levels of the linemod dataset which does not contain occlusions. We note that method of \cite{wohlhart2015learning} was also trained with both clean and occluded images.

\begin{figure}
  \centering
  \includegraphics[width=0.96\linewidth]{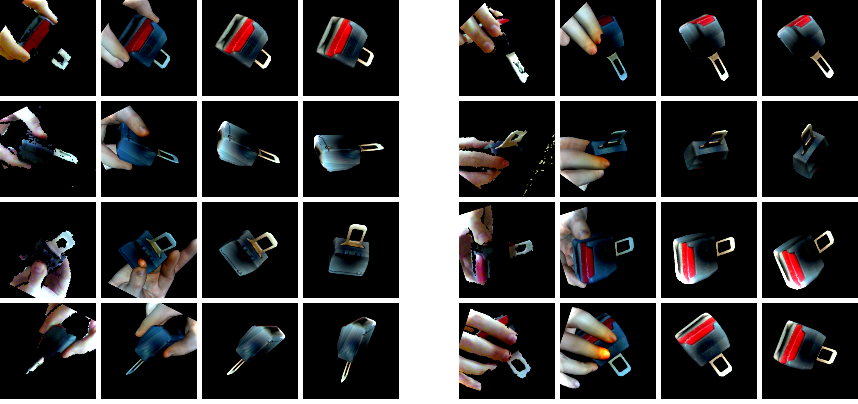}
  \caption{Our occlusion dataset. First column shows a real RGBD image, second column shows the synthetic image rendered using the tracking ground truth annotation, the third column shows the rendered not occluded object corresponding to the occluded image, and the forth column shows our network final estimation.}
  \label{fig:dataset}
\end{figure}

Fig. \ref{fig:dataset} illustrates images and results of our novel hand-object occlusion dataset. From left to right columns represent: a real RGBD image; a synthetic one rendered using our tracker result; the rendered not occluded object that corresponds to the exact pose of the respective occluded real RGBD image; and our network final estimation. 

Regarding our implementation, it was written in Theano. Training one epoch using Nvidia Titan X takes about 15mins for our dataset and about 20 seconds on linemod. One image of our dataset is $96 \times 96$ and takes 4ms for regression and 6ms for NN. Linemod dataset contains images of $64 \times 64$ and evaluation of an images takes about 2ms for regression and 4ms for NN.

\begin{table}[]
\centering
\begin{tabular}{c|c|c|c|c}
\textbf{\begin{tabular}[c]{@{}c@{}}Object\end{tabular}}                                                       & \textbf{\begin{tabular}[c]{@{}c@{}}Nearest\\ Neighbor\\ \cite{wohlhart2015learning}\end{tabular}} & \textbf{\begin{tabular}[c]{@{}c@{}}Siamese\\ Regression\\ Network\\ (End-to-End)\end{tabular}} & \textbf{\begin{tabular}[c]{@{}c@{}}Siamese\\ Regression\\ Features + NN\end{tabular}} & \textbf{\begin{tabular}[c]{@{}c@{}}Siamese\\ Regression\\ + Occlusion\\ Term\end{tabular}} \\ \hline
\textbf{ape}                                                                & 15                                                                               & 12.3                                                                                           & \textbf{11.8}                                                                         & -                                                                                          \\
\textbf{benchviseblue}                                                      & 15.5                                                                             & 15.6                                                                                           & \textbf{13.2}                                                                         & -                                                                                          \\
\textbf{camera}                                                             & 12                                                                               & 10.9                                                                                           & \textbf{10.1}                                                                         & -                                                                                          \\
\textbf{can}                                                                & 15.5                                                                             & 14.5                                                                                           & \textbf{12.3}                                                                         & -                                                                                          \\
\textbf{cat}                                                                & 14                                                                               & 12.1                                                                                           & \textbf{10.4}                                                                         & -                                                                                          \\
\textbf{driller}                                                            & 17.8                                                                             & 16.7                                                                                           & \textbf{13.2}                                                                         & -                                                                                          \\
\textbf{duck}                                                               & 13.9                                                                             & 13.1                                                                                           & \textbf{10.9}                                                                         & -                                                                                          \\
\textbf{holepuncher}                                                        & 13.2                                                                             & 12.9                                                                                           & \textbf{11.4}                                                                         & -                                                                                          \\
\textbf{iron}                                                               & 11.4                                                                             & 11.6                                                                                           & \textbf{10.2}                                                                         & -                                                                                          \\
\textbf{lamp}                                                               & 13.3                                                                             & 12.6                                                                                           & \textbf{11.1}                                                                         & -                                                                                          \\
\textbf{phone}                                                              & 18.2                                                                             & 12.9                                                                                           & \textbf{11.7}                                                                         & -                                                                                          \\ \hline
\textbf{average}                                                              & 14.5                                                                             & 13.2                                                                                           & \textbf{11.4}                                                                         & -                                                                                          \\ \hline
\textbf{\begin{tabular}[c]{@{}c@{}}belt\\ (occlusion dataset)\end{tabular}} & 25.2                                                                             & 13.2                                                                                           & 14.3                                                                                  & \textbf{11.8}                                                                             
\end{tabular}
\vspace{0.05in}
\caption{State of the art and self comparisons of our method against the one of Wohlhart et al. \cite{wohlhart2015learning} in the dataset of LINEMOD \cite{hinterstoisser2012accv} and our novel hand-object.}
\label{soa-table}
\end{table}


\section{Conclusion}\label{Sec_Con}
We presented Siamese Regression Networks, a convolutional network that is able to perform object pose regression in angle space directly, by enforcing distance similarity in feature and pose space among the training samples. Such network is able to learn more discriminative features that are optimal for the pose regression task, which outperform state of the art. Last, our feature-guided pose estimation can be easily modified to learn features that are robust to occlusions, achieving accuracy compared to occlusion free images, when tested on our own severe occlusion-by-hand dataset. As a future work, we would like to investigate how this network can be extended in order to simultaneously tackle object localization, as well as object classification. 


\bibliographystyle{ieee}
\bibliography{main}

\end{document}